\documentclass[letterpaper, 10 pt, conference]{ieeeconf}
\IEEEoverridecommandlockouts                         
\usepackage{algorithmic}
\usepackage[linesnumbered,ruled,vlined]{algorithm2e}
\usepackage{graphics} 
\usepackage{epsfig}
\usepackage{times} 
\usepackage{subfigure}
\usepackage{xcolor}
\usepackage{mathrsfs,graphicx,color,CJK,float,indentfirst,textcomp}
\usepackage{latexsym,amssymb,lscape,amsmath,amsfonts,amssymb,cite,enumerate}
\usepackage{todonotes}
\usepackage{url}
\usepackage{booktabs}
\usepackage{array}
\usepackage{threeparttable}
\usepackage{lipsum}
\usepackage{diagbox}
\usepackage{bm}

\newtheorem{remark}{Remark}[section]

\UseRawInputEncoding

\title{\huge \textbf{Uncertainty-Aware Decision-Making and Planning\\ for Autonomous Forced Merging}}

\author{Jian Zhou\textsuperscript{1}, Yulong Gao\textsuperscript{2}, Bj\"orn Olofsson\textsuperscript{1, 3}, and Erik Frisk\textsuperscript{1}
\thanks{*This research was supported by the Strategic Research Area at Link\"oping-Lund in Information Technology (ELLIIT).}
\thanks{$^{1}$Division of Vehicular Systems, Department of Electrical Engineering, Link\"oping University, Link\"oping, Sweden. $^{2}$Department of Electrical and Electronic Engineering, Imperial College London, London, United Kingdom. $^{3}$Department of Automatic Control, Lund University, Lund, Sweden. Email:
\texttt{jian.zhou@liu.se}, \texttt{yulong.gao@imperial.ac.uk}, \texttt{bjorn.olofsson@control.lth.se}, \texttt{erik.frisk@liu.se}.
}}

\begin{document}
\maketitle
\thispagestyle{empty}
\pagestyle{empty}

\begin{abstract}
In this paper, we develop an uncertainty-aware decision-making and motion-planning method for an autonomous ego vehicle in forced merging scenarios, considering the motion uncertainty of surrounding vehicles. The method dynamically captures the uncertainty of surrounding vehicles by online estimation of their acceleration bounds, enabling a reactive but rapid understanding of the uncertainty characteristics of the surrounding vehicles. By leveraging these estimated bounds, a non-conservative forward occupancy of surrounding vehicles is predicted over a horizon, which is incorporated in both the decision-making process and the motion-planning strategy, to enhance the resilience and safety of the planned reference trajectory. The method successfully fulfills the tasks in challenging forced merging scenarios, and the properties are illustrated by comparison with several alternative approaches. 
\end{abstract}

\section{Introduction}
Motion planning in autonomous driving techniques has demonstrated high potential for saving lives in critical situations \cite{zhou2024homotopic}. However, the motion uncertainties of dynamic obstacles introduce challenges to the safety in motion-planning problems. Among various uncertain scenarios, forced merging on highways presents notable difficulties, as it demands that the autonomous ego vehicle (EV) performs a safety-critical maneuver within a constrained timeframe and space in the presence of dynamic surrounding vehicles (SVs) \cite{fors2022resilient}.

\begin{figure}[!t]
\centering
\includegraphics[width=\columnwidth]{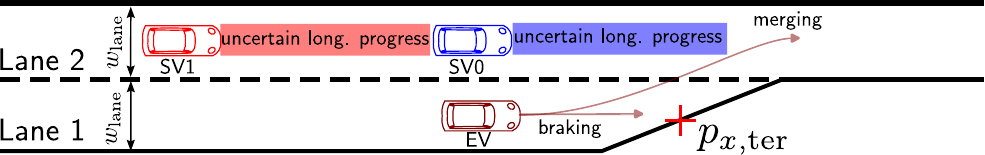}
\caption{The forced merging scenario on the highway. Lane~1 is the current lane of the EV, and lane~2 is the target lane, $p_{x, {\rm ter}}$ is the longitudinal terminal position of lane~1, and $w_{\rm lane}$ is the lane width. Note that more surrounding vehicles can be involved in the scenario.}
\label{fig:merging_scenario}
\end{figure}

The forced merging scenario is shown in Fig.~\ref{fig:merging_scenario}, where the mission of the EV is to merge into lane~2 before reaching $p_{x, {\rm ter}}$ without any collision with SVs and the road boundary, or safely stop in lane 1 before $p_{x, {\rm ter}}$ \cite{liu2022interaction}. The EV needs to first predict the longitudinal progress of the SVs over a horizon and then decide to merge to the target lane or to brake in the current lane. Following the decision, the EV plans a collision-free reference trajectory for merging or braking. The major difficulty in this process is that the motion of the SVs is uncertain, which directly affects the robustness of decision-making and motion planning. One straightforward approach to handling the problem is to consider the worst-case uncertainties of SVs. This method, although reserving sufficient safety margin between the EV and SVs, may result in overly conservative or even infeasible problems \cite{liu2023radius}.

This paper develops an uncertainty-aware decision-making and motion-planning strategy by dynamically estimating the acceleration bounds of each SV. With the estimated acceleration bounds, the forward occupancy of the SVs over the prediction horizon is computed to make a resilient decision at every time step. Following the decision and the SVs' occupancy, the reference trajectory for the EV is calculated in a receding horizon manner by model predictive control (MPC) \cite{zhou2022interactionnew} to fit variations in dynamic traffic environments.

The main contributions of this paper are as follows. (i) The method is aware of the environmental uncertainties using online estimation of the acceleration bounds of the SVs, such that it dynamically captures the uncertainties of SVs without inferring their intentions. (ii) The method achieves twofold resilient planning in forced merging scenarios by integrating the predicted SVs' occupancy in both the decision-making algorithm, which computes a safe reference state at every time step, and the motion-planning strategy, which plans a safe trajectory by tracking the reference state. (iii) The effectiveness of the method is verified in comparison with alternative approaches in simulations with the SVs simulated based on a recorded traffic dataset from the real world.

\subsection{Related Work}\label{sec: related workd}
A wide range of approaches for uncertainty-aware motion planning has been explored, focusing primarily on predicting the uncertain intentions of SVs or the occupancy of SVs over a given horizon. This review will cover related work on these two methodological categories in the context of merging scenarios. These methods are also applicable to other highway driving situations with certain adaptations.

\paragraph*{\textbf{Intention-Prediction Based Planning}} This type of method characterizes the SVs' uncertainties as finite multi-modal intentions, inferred based on factors like interactions with the traffic environment. The control policy for the EV is optimized corresponding to the multi-modal uncertainty predictions. As the EV's policies can represent different maneuvers, such as lane changing, stopping, or re-entry, the optimization simultaneously facilitates decision-making and trajectory planning. For example, \cite{liu2022interaction} proposed a Leader-Follower Game to predict the intentions of the SVs, which were assumed to perform cooperative behaviors with the EV in the merging scenario, and then a stochastic MPC was designed for the EV to find the best policy by optimizing over finite motion primitives. In \cite{ulfsjoo2022integrating}, the probabilities of SVs' intentions were predicted by a classifier trained by a traffic dataset, then the decision-making and motion planning were achieved by integrating a partially observable Markov decision process with a scenario MPC. In \cite{chen2022interactive}, a branch MPC was designed for uncertainty-aware motion planning of an EV, where the uncertain maneuvers of an SV were described by finite policies with known distributions. Similar methods were investigated in \cite{wang2023interaction} and \cite{oliveira2023interaction} that identified the distributions of policies of an SV, which was assumed to adhere to safety considerations, in real-time. The prediction was then utilized to construct an MPC-based planner for merging. In \cite{fors2022resilient}, the joint decision-making and motion planning was made by expanding a scenario tree for the EV, while the SVs were assumed to drive with adversarial intentions to the EV.

\paragraph*{{\textbf{Occupancy-Prediction Based Planning}}} This kind of method computes either an occupancy that covers all reachable positions of each SV, or a subset of the full occupancy area with a reduction of conservativeness. For instance, \cite{xu2014motion} and \cite{suh2018stochastic} predicted the SV's occupancy in the merging scenario using the Kalman filter and the extended Kalman filter, respectively, by approximating uncertainties with a normal distribution, to design a motion planner. In \cite{brudigam2023stochastic}, by assuming that the uncertainties of SVs were subject to a Gaussian distribution, the SVs were predicted with multi-modal uncertainties to design a safety-guaranteed stochastic MPC planner. In \cite{liu2023radius}, a risk-bounded reachability-based planning strategy was proposed for both merging and intersection scenarios, where the SVs were predicted based on an assumed tractable probability density function. In \cite{yoon2021interaction}, the motion of SVs in the cut-in scenario was predicted by a trained Gaussian process model, and the uncertainty was propagated by an extended Kalman filter for the subsequent design of an MPC-based planner. The most closely related work to the proposed method is in \cite{zhou2024robust}, which estimated the control set for obstacles by solving a linear programming problem. This differs from this paper in two major technical aspects. Firstly, instead of solving an optimization problem to learn the control set in \cite{zhou2024robust}, this paper can analytically estimate the control set of SVs in the forced merging.  Secondly, compared with executing a predefined decision of the EV in \cite{zhou2024robust}, this research integrates the uncertainty prediction of SVs into both the decision-making and the motion-planning strategies to enhance the resilience of the motion-planning method. 

\section{Notations and Problem Statement}\label{sec: Problem Statement}
This section introduces the general notations of the paper, then presents the motion models of EV and SVs in Fig.~\ref{fig:merging_scenario}, and finally states the specific research problems to be solved.

\subsubsection{Notations} $\mathbb{R}^n$ is the $n$-dimensional real number vector space, $\mathbb{R}_{+}^n$ is the $n$-dimensional non-negative vector space, and $\mathbb{N}_{+}^n$ is the $n$-dimensional positive integer vector space. $I^n$ indicates an $n \times n$ identity matrix. For two sets $\mathcal{A}$ and $\mathcal{B}$, $\mathcal{A} \oplus \mathcal{B} = \left\{ x + y \ | \ x \in \mathcal{A}, y \in \mathcal{B} \right\}$. For a set $\Gamma \subset \mathbb{R}^{n_a+n_b}$, the set projection is defined as $\mathcal{A} = {\rm Proj}_a(\Gamma)=\left\{ a\in \mathbb{R}^{n_a}: \exists b\in \mathbb{R}^{n_b}, (a, b)\in \Gamma \right\}$.  An interval of integers is denoted by $\mathbb{I}_a^b = \{a, a+1, \cdots, b\}$. A matrix of appropriate dimension with all elements equal to 0 is denoted by ${\bm 0}$. The current time step is indicated by $t$, and the prediction of a variable $x$ at time step $t+i$ over the prediction horizon is represented as $x_{i|t}$. 

\subsubsection{Modeling of Vehicles}\label{sec: modeling vehicles}
In the merging scenario, the EV is modeled by a single-track kinematic model \cite{dixit2020trajectory} as 
\begin{subequations}\label{eq: EV model}
\begin{align}
(\dot{p}_x^e, \ \dot{p}_y^e) & = \left(v^e, \ v^e\left(\varphi^e+\frac{l_r}{l_f + l_r}\delta^e\right)\right), \label{eq_EV_model_a} \\
(\dot{\varphi}^e, \ \dot{v}^e, \ \dot{a}^e)  & = \left(\frac{v^e\delta^e}{l_f+l_r}, \ a^e, \ \eta^e\right), \label{eq_EV_model_c}  
\end{align}
\end{subequations}
\noindent where $(p_x^e, p_y^e)$ is the coordinate of the center of geometry in the ground coordinate system. The variable $\varphi^e$ is the inertial yaw angle, $v^e$, $a^e$, and $\eta^e$ mean the longitudinal speed, acceleration, and jerk in the vehicle frame, respectively, $\delta^e$ is the front wheel angle, and $l_f$ and $l_r$ are the distances from the center of geometry to the front axle and rear axle, respectively. The model~\eqref{eq: EV model} is linear in $\varphi^e$, $\delta^e$, as they are assumed small during the merging process. This approximation is beneficial for computational efficiency. The discrete-time form of the model \eqref{eq: EV model} is
\begin{equation}
x_{t+1}^e = f(x_{t}^e, u_{t}^e), \label{eq: EV model RK}	
\end{equation} 
\noindent where $x_{t}^e = [p_{x, {t}}^e \ p_{y,{t}}^e \ \varphi_{t}^e \ v_{t}^e \ a_{t}^e]^{\top}$ is the state, and $u_{t}^e = [\delta_{t}^e \ \eta_{t}^e]^{\top}$ is the control. Eq.~\eqref{eq: EV model RK} is obtained by a fourth-order Runge-Kutta method with the fixed step-length $T$.

The SVs in Fig.~\ref{fig:merging_scenario} have no lateral motion since they keep in lane 2, while the longitudinal accelerations are uncertain. Denote by $s \in \mathcal{S}$ the index of the SV, where $\mathcal{S}$ is the index set of the SVs. In this case, the motion of the center of geometry of SV $s$ can be modeled by a linear model as
\begin{equation}
x_{t+1}^s = A^sx_{t}^s + B^sa_{x,t}^s,\label{eq: SV model}
\end{equation} 
\noindent where  $A^s = \begin{bmatrix} 1 & T \\ 0 & 1 \end{bmatrix}$ and $B^s = \begin{bmatrix} T^2/2 \\ T \end{bmatrix}$ are system matrices, and $T$ is the sampling interval. The state vector $x_t^s = [p_{x,t}^s \ v_{x,t}^s]^{\top}$ contains the longitudinal position and velocity of SV $s$ at time step $t$ in the ground coordinate system. The variable $a_{x, t}^s $ means the longitudinal acceleration (control input) of the SV at time step $t$ in the ground coordinate system. Note that $a_{x, t}^s $ is the same as the acceleration of the SV in the vehicle frame in Fig.~\ref{fig:merging_scenario}. The control input of the SV is unknown yet bounded, and the worst-case acceleration set covering all possible control actions of the SV $s$ can be represented as\cite{fors2021predictive}
\begin{equation}
\mathbb{U}^s=\{a: -\mu g \leq a \leq \mu g\},\label{eq: SV worst bound}
\end{equation} 
where  $\mu$ is the road-friction coefficient, and $g$ is the constant of acceleration.

\subsection{Research Problems}\label{sec: research problem}
Given a prediction horizon from the current time step $t$, the control inputs of model~\eqref{eq: SV model} of the SVs are uncertain over the horizon. However, considering the worst-case set $\mathbb{U}^s$ of the SVs can be overly conservative for the EV's planning. Therefore, a resilient decision-making and motion-planning strategy depends on real-time estimation of the acceleration bounds of SVs to pursue robustness and efficiency. Following this idea, the research problems can be specified as three subproblems to be solved. (i) Estimate the acceleration bound of model~\eqref{eq: SV model} of each SV using online data. (ii) Predict the forward occupancy of the SVs based on model~\eqref{eq: SV model} using the estimated acceleration bounds. (iii) Design the uncertainty-aware decision-making and motion-planning strategy for the EV model~\eqref{eq: EV model RK} based on the predicted occupancy of SVs. The first two subproblems are addressed in Section~\ref{sec: Uncertainty Prediction of Surrounding Vehicles}, and the last subproblem is solved in Section~\ref{sec: Robust Decision-Making and Planning}.

\section{Uncertainty Predictions}\label{sec: Uncertainty Prediction of Surrounding Vehicles}
This section introduces the methods for estimating the acceleration bounds of SVs and predicting the forward occupancy of the SVs based on the SVs' model~\eqref{eq: SV model}.

\subsection{Estimating Acceleration Bounds of Surrounding Vehicles}\label{sec: Estimating the Input Bound of Surrounding Vehicles}
The accelerations of the SVs are affected by many factors like the traffic environment, the tire--road interactions, the driver's preference, etc., which make the acceleration bounds rather complex to predict. Considering that the behaviors of a system can be reflected through historical data\cite{hu2023active}, \cite{gillula2013reducing}, this paper proposes an efficient way to infer the bound based on an information set containing observed accelerations of the SVs until the current time step $t$. The information set for the SV $s$ at time step $t$ is defined as
\begin{equation}\label{eq: information set}
\mathcal{I}_t^s = \{a_{x,k}^s: k \in \mathbb{I}_0^{t-1}\}\subseteq \mathbb{U}^s.
\end{equation}

Based on the set $\mathcal{I}_t^s$, the estimated acceleration set $\hat{\mathbb{U}}_0^s$ of the SV~$s$ at the initial time step $t=0$ can be represented as
\begin{equation}\label{eq: U_hat}
\hat{\mathbb{U}}_0^s = \left\{a:  a_{x,0}^{s, \mathrm{min}}\leq a \leq a_{x,0}^{s, \mathrm{max}}\right\},
\end{equation}
\noindent where $a_{x,0}^{s, \mathrm{min}}= {\rm min}\{\mathcal{I}_0^s\}$ and $a_{x,0}^{s, \mathrm{max}}= {\rm max}\{\mathcal{I}_0^s\}$. The initial information set $\mathcal{I}_0^s$ can be constructed as a non-empty set containing selected elements or the observed accelerations of the SV $s$ before time step $0$. Then, for $t\in \mathbb{N}_+$, the bounds $a_{x,t}^{s, \rm min}$ and $a_{x,t}^{s, \rm max}$ are recursively updated as
\begin{subequations}\label{eq: update acc bound online}
\begin{align}
a_{x,t}^{s, \mathrm{min}} &= \min\{ a_{x,t-1}^{s, \mathrm{min}}, a_{x,t-1}^s\}, \\
a_{x,t}^{s, \mathrm{max}} &= \max\{ a_{x,t-1}^{s, \mathrm{max}}, a_{x,t-1}^s\},
\end{align}	
\end{subequations}
\noindent where $a_{x,t-1}^s$ is the observed acceleration of SV $s$ at time step $t-1$. The bounds $a_{x,t}^{s, \rm min}$ and $a_{x,t}^{s, \rm max}$ construct the estimated acceleration set of the SV, which is represented as $\hat{\mathbb{U}}_t^s$, in the form of \eqref{eq: U_hat}. The set $\hat{\mathbb{U}}_t^s$ is then used to predict the forward occupancy of the SV in Section~\ref{sec: Forward Occupancy Prediction}. Note that the formulation~\eqref{eq: update acc bound online} can rapidly capture the change of driving behaviors of SVs when a control signal that represents the change of driving style of the SV is observed. In addition, since $a_{x,t}^s$ is one dimensional, \eqref{eq: update acc bound online} is more efficient than the approach in  \cite{zhou2024robust} that estimates the control set of the obstacles by solving a scenario-optimization problem. 

\subsection{Predicting the Forward Occupancy}\label{sec: Forward Occupancy Prediction}
\begin{figure}[!t]
\centering
\includegraphics[width=0.65\columnwidth]{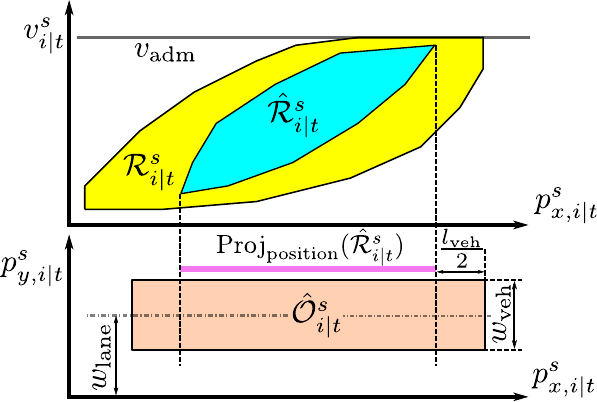}
\caption{The predicted reachable set and occupancy of the SV.}
\label{fig:visualized_reachable_set_project}
\end{figure}
The forward occupancy of SV $s$, denoted by $\hat{\mathcal{O}}_{i|t}^s$, defines a region that might be occupied by the agent at time step $t+i$ in the prediction horizon. The occupancy $\hat{\mathcal{O}}_{i|t}^s$ is calculated through forward reachability analysis as \cite{gao2022risk}
\begin{subequations}\label{eq: forward reach} 
\begin{align}
\hat{\mathcal{O}}_{i|t}^s & = {\rm Proj}_{\rm position}(\hat{\mathcal{R}}_{i|t}^s) \oplus \mathcal{O}_t^s, \label{eq: occupancy prediction} \\
\hat{\mathcal{R}}^s_{i|t} & = \left(A^s \hat{\mathcal{R}}^s_{i-1|t} \oplus B^s \hat{\mathbb{U}}_t^s \right) \bigcap \mathbb{X}^s, \label{eq: reachable set prediction},\\
\mathbb{X}^s&=\left\{x: [0 \ 0]^{\top} \leq x \leq [+\infty \ v_{\rm adm}]^{\top}\right\}, \label{eq: SV state admissible}
\end{align}
\end{subequations}
\noindent where $A^s$, $B^s$ are defined in \eqref{eq: SV model}, $\mathbb{X}^s$ is the state admissible set of system~\eqref{eq: SV model}, and $v_{\rm adm}$ is the upper bound of the admissible speed of the SVs. The set $\hat{\mathcal{R}}_{i|t}^s$ is the reachable set of the full state of the SV predicted at time step $t+i$ in the prediction horizon. The forward reachability prediction is initiated with $\hat{\mathcal{R}}_{0|t}^s = \left\{x_t^s \right\}$, where $x_t^s$ is the measured state of the SV~$s$ at time step $t$. The operation ${\rm Proj}_{\rm position}(\hat{\mathcal{R}}_{i|t}^s)$ means projecting the set of positions from the full reachable set. The occupancy of the SV at the current time step $t$ is
$$\mathcal{O}_t^s = \left\{x: -\left[l_{\rm veh}/2 \ w_{\rm veh}/2\right]^{\top} \leq x \leq \left[l_{\rm veh}/2 \ w_{\rm veh}/2\right]^{\top}\right\},$$
\noindent where $l_{\rm veh}$ and $w_{\rm veh}$ mean the vehicle length and width, respectively, which are assumed equal for the EV and SVs. 

Replacing $\hat{\mathbb{U}}_t^s$ by $\mathbb{U}^s$, which is defined in \eqref{eq: SV worst bound}, in \eqref{eq: reachable set prediction} yields the full reachable set $\mathcal{R}_{i|t}^s$. The difference between $\hat{\mathcal{R}}_{i|t}^s$ and $\mathcal{R}_{i|t}^s$ is shown in Fig.~\ref{fig:visualized_reachable_set_project}. It is seen that $\hat{\mathcal{R}}_{i|t}^s \subseteq \mathcal{R}_{i|t}^s$, such that it can reduce the conservativeness in the decision-making and motion-planning strategy as detailed in Section~\ref{sec: Robust Decision-Making and Planning}. 

\section{Uncertainty-Aware Decision-Making and Motion Planning for the Ego Vehicle}\label{sec: Robust Decision-Making and Planning}
Following the forward occupancy $\hat{\mathcal{O}}_{i|t}^s$ predicted in~\eqref{eq: forward reach}, this section introduces the uncertainty-aware decision-making and motion-planning strategy for the EV in the forced merging scenario in the presence of uncertain SVs. 

\subsection{Decision-Making Strategy}\label{sec: decision-making strategy}
\begin{algorithm}[t!]\small
    \renewcommand{\thealgocf}{I}
    \DontPrintSemicolon
    \KwIn{$v_{x, t}^e$, $p_{x, t}^e$, $p_{y, t}^e$, $p_{x, t}^s$, $\hat{\mathcal{O}}_{i|t}^s$, $\forall s \in \left\{0, 1\right\}$, $i=1, \ldots, N$.}
    \KwOut{$v_{x, t}^m$: Reference velocity of maneuver $m$.}
    $\underline{p}_{x, i|t} = -\infty, \ \overline{p}_{x, i|t} = p_{x, \rm ter}$, $v_{x, t}^{{\rm VT1}} = {\rm argmin}\left\{{\rm QP}\right\}$. \\
    $\overline{p}_{x, i|t}^s = {\rm max}\left\{x \ | \ \exists y, (x, y) \in \hat{\mathcal{O}}_{i|t}^s\right\}$. \\
    $\underline{p}_{x, i|t}^s = {\rm min}\left\{x \ | \ \exists y, (x, y) \in \hat{\mathcal{O}}_{i|t}^s\right\}$. \\
    \If{$p_{x, t}^{0} \leq p_{x, t}^e$}
    {let $\underline{p}_{x, i|t} = \overline{p}_{x, i|t}^0, \ \overline{p}_{x, i|t} = +\infty$, \\
    $v_{x, t}^{{\rm VT2}} = {\rm argmin}\left\{{\rm QP}\right\}$.}
    \ElseIf{$p_{x, t}^{1} \leq p_{x, t}^e < p_{x, t}^0$}
    {let $\underline{p}_{x, i|t} = \overline{p}_{x, i|t}^1, \ \overline{p}_{x, i|t} = \underline{p}_{x, i|t}^0$, \\
    \If{${\rm min}\left\{\overline{p}_{x, i|t} - \underline{p}_{x, i|t}\right\} \leq 2d_{\rm min}^{\rm qp}$}
    {$v_{x,t}^{\rm VT2} = 0$}
    \Else{$v_{x, t}^{{\rm VT2}} = {\rm argmin}\left\{{\rm QP}\right\}$.}}
    \Else
    {let $\underline{p}_{x, i|t} = -\infty, \ \overline{p}_{x, i|t} = \underline{p}_{x, i|t}^1$, \\
    $v_{x, t}^{{\rm VT2}} = {\rm argmin}\left\{{\rm QP}\right\}$}
    \caption{\small{Computing reference velocities of the EV}}
    \label{alg:ref_vel}
    \end{algorithm}

In the scenario shown in Fig.~\ref{fig:merging_scenario}, the decision-making strategy computes the reference state for the EV at every time step $t$. The reference state is defined as $[v_{x, t}^{\rm ref} \ y_{x, t}^{\rm ref}]$, containing the terminal reference velocity and the lateral reference position. They are computed following three steps.

\textbf{Step 1}: Define the maneuver set of the EV.

The maneuver set of the EV, denoted by $\mathcal{M}^e$, includes the candidate maneuvers for the EV. Since the road in Fig.~\ref{fig:merging_scenario} has two lanes for the forced merging scenario, the set is designed as $\mathcal{M}^e= \left\{\text{VT1, VT2}\right\}$, where VT1 and VT2 mean velocity-tracking maneuver on lane 1 and lane 2, respectively. The velocity-tracking mode tracks a reference velocity in the current lane, which is calculated in Step 2 (the velocity-tracking model is defined in \eqref{Eq: QP d}).

\textbf{Step 2}: Compute the reference of each maneuver.
The lateral reference position and longitudinal reference velocity for maneuver $m \in \mathcal{M}^e$ at time step $t$ are denoted as $p^{m}_{y, t}$ and $v^m_{x, t}$, respectively. The lateral reference position is the center of the target lane, i.e., for $m = {\text{VT1}}$, $p^m_{y, t} = 0.5w_{\rm lane}$, and for $m = {\text{VT2}}$, $p^m_{y, t} = 1.5w_{\rm lane}$, where $w_{\rm lane}$ is the lane width. The safe longitudinal reference velocity is computed by Algorithm~\ref{alg:ref_vel}. In Algorithm~\ref{alg:ref_vel}, $v_{x, t}^e$, $(p_{x, t}^e, p_{y, t}^e)$, $p_{x, t}^{\rm 0}$, and $p_{x, t}^{\rm 1}$ denote the longitudinal velocity of the EV, the coordinate of the geometric center of the EV, and the longitudinal geometric center of SV0 and SV1 (see Fig.~\ref{fig:merging_scenario}) at time step $t$ in the ground coordinate system, respectively. The parameterized quadratic-programming (QP) problem at lines 1, 6, 12, and 15 in Algorithm~\ref{alg:ref_vel} is defined as \cite{zhou2022interactionnew} and \cite{zhou2022interaction}
\begin{subequations}\label{Eq:QP}
    \begin{align}
    \mathop{\rm minimize}\limits_{v_x^{{\rm ref}}} \quad & ||v_{x, i|t}^e - v_x^{{\rm ref}}||_2^2 & \label{Eq: QP a} \\
    {\rm subject \ to}\quad
    &[p_{x,i|t}^e \ v_{x, i|t}^e]^{\top} = Hz_{i|t}^e,& \label{Eq: QP b}\\
    &\underline{p}_{x,i|t} + d_{\rm min}^{\rm qp} \leq p_{x,i|t}^{e} \leq \overline{p}_{x,i|t} - d_{\rm min}^{\rm qp}, & \label{Eq: QP c}\\
    &z_{i|t}^{e} = \Phi^e z_{i-1|t}^{e} + B^eK^ez^{{\rm r}} , & \label{Eq: QP d}
    \end{align}
    \end{subequations}
\noindent where $i=1, \ldots, N$, and $N$ is the prediction horizon in the motion prediction. The cost function~\eqref{Eq: QP a} minimizes the deviation between $v_x^{\rm ref}$, the target velocity, and $v_{x, i|t}^e$, the predicted longitudinal velocity of the EV at time step $t+i$ in the ground coordinate system. In~\eqref{Eq: QP b}, $p_{x,i|t}^e$ is the predicted longitudinal position of the EV at time step $t+i$. The matrix $H$ selects the elements $p_{x,i|t}^e$ and $v_{x, i|t}^e$ from the full state $z_{i|t}^{e}=[p_{x, i|t}^e \ v_{x, i|t}^e \ a_{x, i|t}^e \ p_{y,i|t}^e \ v_{y, i|t}^e \ a_{y, i|t}^e]^{\top}$, which contains the measured longitudinal position, velocity, acceleration, and lateral position, velocity, acceleration of the EV in the ground coordinate system at time step $t + i$. In~\eqref{Eq: QP c}, the parameters $\underline{p}_{x,i|t}$ and $\overline{p}_{x,i|t}$ are lower and upper bounds on the EV's longitudinal position, respectively. The parameter $d_{\rm min}^{\rm qp}$ is defined as $d_{\rm min}^{\rm qp} = d_{\rm min} + l_{\rm veh}$, where $d_{\rm min}$ is the minimum safety distance. Constraint~\eqref{Eq: QP d} defines the model for predicting the state $z_{i|t}^e$ by a point-mass linear state-feedback model \cite{lefkopoulos2021interaction}, with the matrices defined as:
\begin{subequations}
\begin{align*}
    \Phi^e &= (A^e-B^eK^e), \\
    A^e &= \left[\begin{matrix}A_x & {\bm 0} \\ {\bm 0} & A_y \end{matrix}\right], \ A_x = A_y = \left[\begin{matrix} 1 & T & T^2/2 \\ 0 & 1 & T \\ 0 & 0 & 1 \end{matrix}\right]\\
B^e &= \left[\begin{matrix}B_x & {\bm 0} \\ {\bm 0} & B_y \end{matrix}\right], \ B_x = \left[\begin{matrix} 0 \\ T^2/2 \\ T \end{matrix}
\right], \ B_y = \left[\begin{matrix} T^3/6 \\ T^2/2 \\ T \end{matrix}\right]\\
K^e &= \left[\begin{matrix}K_{\rm lon}^{\top} & {\bm 0} \\ {\bm 0} & K_{\rm lat}^{\top} \end{matrix}\right], \ K_{\rm lon} \in \mathbb{R}^3, \ K_{\rm lat} \in \mathbb{R}^3.
\end{align*}
\end{subequations}

The reference state in~\eqref{Eq: QP d} is $z^{{\rm r}} = [0 \ v_x^{\rm ref} \ 0 \ p_y^{\rm ref} \ 0 \ 0]^{\top}$, where $v_x^{\rm ref}$ is the optimization variable in \eqref{Eq:QP}, $p_y^{\rm ref} = 0.5w_{\rm lane}$ at line~1 in Algorithm~\ref{alg:ref_vel}, and $p_y^{\rm ref} = 1.5w_{\rm lane}$ at line~6, line~12, and line~15 in Algorithm~\ref{alg:ref_vel}. 

\textbf{Step 3}: Decide the EV's optimal maneuver at time step $t$.

The cost of maneuver $m$ with $v_{x, t}^m$ and $p_{y, t}^m$ is~\cite{zhou2022interactionnew}
\begin{equation*}
\begin{split}
J_t^m = & \sum_{i=1}^{N}\left(\left\|a_{x, i|t}^{e,m}\right\|_{W_x}^2 + \left\|a_{y,i|t}^{e,m}\right\|_{W_y}^2\right) + \\
&\left\|v_{x, t}^{e} - v_{x, t}^m \right\|_{W_v}^2 +
\left\|p_{y, t}^{e} - p_{y, t}^m\right\|_{W_l}^2, \label{eq_model_cost}
\end{split}
\end{equation*} 
\noindent where $W_x$, $W_y$, $W_v$, and $W_l$ are weights. The cost is used to decide the reference maneuver of the EV at time step $t$, which is denoted by $m_t^e$,  and is chosen as
\begin{equation*}
m_t^e = \mathop{\arg\max}\limits_{m \in \mathcal{M}^e} \ \left\{P(m) \bigg| P(m)=\frac{1/{\sqrt{J_t^m}}}{\sum_{q \in \mathcal{M}_e}{1}/{\sqrt{J_t^q}}}\right\}. \label{eq_select_model}
\end{equation*}

The references associated with maneuver $m_t^e$ are $[v_{x, t}^{\rm ref} \ p_{y, t}^{\rm ref}]$. They are called moving targets because they are updated at every time step to enhance the collision-avoidance ability compared with tracking a constant reference \cite{zhou2022interaction}.

\subsection{Motion-Planning Strategy}\label{sec: MPC Design for Merging}
The motion planner by MPC is designed to track the reference $[v_{x, t}^{\rm ref} \ y_{x, t}^{\rm ref}]$ subject to collision-avoidance constraints with SVs, and the differential constraints of the EV model. The MPC optimization problem at time step $t$ is
\begin{subequations}
\begin{align}
\mathop{\rm minimize}\limits_{u_{i-1|t}^e, \ \lambda_{i}^s} \quad & \sum_{i=1}^{N_p}\left(||\delta_{i-1|t}^e||_{Q_1}^2 +
|| \eta_{i-1|t}^e||_{Q_2}^2\right) + ||E_{N_p|t}||_{Q_3}^2 & \label{eq_speot_ocp_a} \\
{\rm subject \ to}\quad
&x_{i|t}^e = f(x_{i-1|t}^e, u_{i-1|t}^e), & \label{eq_speot_ocp_b}\\
&\underline{\mathcal{U}} \leq \left[v_{i|t}^e \ a_{i|t}^e \ \delta_{i-1|t}^e\right]^{\top} \leq \overline{\mathcal{U}}, &\label{eq_speot_ocp_c}\\
& {p_{i|t}^e} \in \mathcal{D},& \label{eq_speot_ocp_d} \\
&(H_{i|t}^sp_{i|t}^e - h_{i|t}^s)^{\top}\lambda_i^s  \geq d_{\rm min}, &\label{eq_speot_ocp_e}\\
&||(H_{i|t}^s)^{\top}\lambda_i^s||_2  \leq 1, \label{eq_speot_ocp_f} \\
&\lambda_i^s  \in \mathbb{R}_{+}^4, \ s \in \mathcal{S},\label{eq_speot_ocp_g}
\end{align}
\end{subequations}
\noindent where $i=1, \ldots, N_p$, and $N_p$ is the prediction horizon in the MPC problem. The vector $E_{N_p|t} = [p_{y, N_p|t}^e-p_{y,t}^{\rm ref} \ v_{N_p|t}^e - v_{x, t}^{\rm ref}]$ describes the terminal deviation. The weighting matrices of the loss function are denoted by $Q_1, \ldots, Q_3$, and constraint~\eqref{eq_speot_ocp_b}, which is initialized with $x_{0|t}^e$, enforces the vehicle model~\eqref{eq: EV model RK}. Constraint~\eqref{eq_speot_ocp_c} provides the bounds on the acceleration, velocity, and steering input of the EV, and \eqref{eq_speot_ocp_d} constrains the position vector of the EV, $p_{i|t}^e = [p_{x, i|t}^e \ p_{y, i|t}^e]^{\top}$, within the drivable area $\mathcal{D}$ over the prediction horizon. Eqs.~\eqref{eq_speot_ocp_e}--\eqref{eq_speot_ocp_g} are equivalent with collision-avoidance constraints between ${p_{i|t}^e}$, the position of the EV, and $\hat{\mathcal{O}}_{i|t}^s$, the predicted occupancy of SV $s$, as below \cite{zhang2021optimization}
\begin{subequations}\label{eq: distance constraint}
\begin{align}
d_{\rm min} & \leq {\rm dist}\left(p_{i|t}^e, \left\{\hat{\mathcal{O}}_{i|t}^s \oplus \mathcal{O}^e\right\}\right), \label{eq_distance_measure} \\
\left\{\hat{\mathcal{O}}_{i|t}^s \oplus \mathcal{O}^e\right\} & = \left\{x \in \mathbb{R}^2: H_{i|t}^s x \leq h_{i|t}^s\right\}, \label{eq_extend_reachable_set}
\end{align}
\end{subequations}
\noindent where $\mathcal{O}^e$ is the EV's occupancy with the geometric center and the heading as zero and ${\rm dist}(\cdot)$ is the distance measure~\cite{yu2024continuous}. Note that the impact of the heading angle of the EV is not considered in \eqref{eq: distance constraint} since the heading angle of the vehicle in the lane-changing scenario is usually small \cite{dixit2020trajectory}. 
\begin{remark}
Algorithm~\ref{alg:ref_vel} is a rule-based reactive decision-making strategy, which has not been studied in \cite{zhou2022interactionnew} and \cite{zhou2022interaction}. The indices of SVs in Algorithm~\ref{alg:ref_vel} are $0$ and $1$ as Algorithm~\ref{alg:ref_vel} is specified for the scenario in Fig.~\ref{fig:merging_scenario}. It can be naturally extended to scenarios with multiple SVs, and the algorithm's complexity does not change as it only solves one QP problem regardless of the number of SVs. In addition, it ensures that, if \eqref{Eq:QP} at line~1 is feasible, the EV can return to lane 1 in a critical situation, e.g., if the SV0 is accelerating when the EV is merging in front of it. 
\end{remark}

\section{Results and Discussion} \label{sec: simulation and discussion}
The performance of the proposed method is evaluated in the merging scenario shown in Fig.~\ref{fig:merging_scenario} in several case studies. The first case study compares the proposed method with a robust MPC (RMPC) and a deterministic MPC (DMPC) to demonstrate the performance of the  proposed method, and the implementations of the RMPC and DMPC will be introduced in Section~\ref{sec: compare with R and D}. The second case study compares the proposed method with an interaction-aware branch MPC, which has been widely investigated recently (see Section~\ref{sec: related workd}), to show how the proposed method performs safe planning without accurate prediction of the intentions of the SVs. The third case study analyzes the convergence of the proposed method with the information set that contains the observed accelerations of the SVs. 

In the case studies, the accelerations of SVs are generated from predefined sets, with support and distribution unknown to the EV. The distributions of the SVs' accelerations are designed based on a recorded traffic dataset, the \texttt{highD} dataset \cite{krajewski2018highd}, to verify the performance of the method in the presence of human-like drivers in the real world. The polytope computations were implemented by the Python package \texttt{pytope} \cite{pytope}. The optimization problems were solved by \texttt{CasADi} \cite{andersson2019casadi} and \texttt{Ipopt} \cite{wachter2006implementation} using the linear solver \texttt{MA57} \cite{hsl2021collection}. All simulation parameters are summarized in Table~\ref{Table_simulatioNarameters forced merging}, and the initial conditions of the vehicles are summarized in Table~\ref{Table_initial condition}. The implementations are found in our published code\footnote{\textcolor{blue}{{\url{https://github.com/JianZhou1212/auto-merging}}}}, and videos of the simulation are available online\footnote{\textcolor{blue}{\url{https://youtu.be/gN9OPjjtnkI}}}. Note that the value of $d_{\rm min}$ is designed relatively small in Table~\ref{Table_simulatioNarameters forced merging} to highlight the performance evaluation of the proposed method and the alternative approaches in safety-critical scenarios. In practical applications, $d_{\rm min}$ can be designed larger to increase the distance to the SVs.

\begin{table}[!t]\tiny
\centering
\caption{Parameters of Forced Merging Simulations}  
\label{Table_simulatioNarameters forced merging} 
\begin{tabular}{cccc}
\toprule
\textbf{Symbol} & \textbf{Value} & \textbf{Symbol} & \textbf{Value} \\
\midrule
$l_f$, $l_r$ & $1.65 \ {\rm m}$, $1.65 \ {\rm m}$ & $T$& $0.25 \ {\rm s}$ \\ \specialrule{0em}{1pt}{1pt}
$d_{\rm min}$ & $0.5 \ {\rm m}$ in \eqref{Eq:QP}, $0.1 \ {\rm m}$ in \eqref{eq_speot_ocp_e} & $l_{\rm veh}$, $v_{\rm veh}$ & $4.3 \ {\rm m}$, $1.8 \ {\rm m}$ \\ \specialrule{0em}{1pt}{1pt}
$W_x, W_y, W_v, W_l$ & 0.1, 0.1, 0.7, 0.1 & $\mu$, $g$& $0.71$, $9.8 \ {\rm m/s^2}$ \\ \specialrule{0em}{1pt}{1pt}
$K_{\rm lon}$ &  $[0 \ 0.3847 \ 0.8663]^{\top}$ & $w_{\rm lane}$ & $4 \ {\rm m}$\\ \specialrule{0em}{1pt}{1pt}
$K_{\rm lat}$& $[0.5681 \ 1.4003 \ 1.7260]^{\top}$ & $v_{\rm adm}$ & $50 \ {\rm m/s}$\\ \specialrule{0em}{1pt}{1pt}
$\underline{\mathcal{U}}$ & $[0  \ -5 \ -0.1]^{\top}$ & $\overline{\mathcal{U}}$& $[50  \ 2.5 \ 0.1]^{\top}$\\ \specialrule{0em}{1pt}{1pt}
$Q_1, Q_2$ & $100$, $0.001$ & $Q_3$ & $\rm{diag}([1 \quad 1])$ \\ \specialrule{0em}{1pt}{1pt}
$N_p$ & $10$ & $N$ & $20$ \\ \specialrule{0em}{1pt}{1pt}
\bottomrule
\end{tabular}
\begin{tablenotes}\tiny
\item ${\star}$ The units in $\overline{\mathcal{U}}$ and $\underline{\mathcal{U}}$ are ${\rm m/s}$, ${\rm m/s^2}$, and ${\rm rad}$, respectively.
\end{tablenotes} 
\end{table}
\begin{table}[!t]\tiny
\centering
\caption{Initial Conditions of the Vehicles in Each Case}  
\label{Table_initial condition} 
\begin{tabular}{cccc}
\toprule
\textbf{Section} & \textbf{EV} & \textbf{SV0} & \textbf{SV1} \\
\midrule
Section~\ref{sec: compare with R and D}, Section~\ref{sec: convergence} & ($822.5 \ {\rm m}$, $30 \ {\rm m/s}$) & ($812.5 \ {\rm m}$, $30 \ {\rm m/s}$) &($772.5 \ {\rm m}$, $30 \ {\rm m/s}$)\\ \specialrule{0em}{1pt}{1pt}
Section~\ref{sec: compare with branch MPC} & ($822.5 \ {\rm m}$, $30 \ {\rm m/s}$) & ($815 \ {\rm m}$, $30 \ {\rm m/s}$) &($772.5 \ {\rm m}$, $30 \ {\rm m/s}$)\\ \specialrule{0em}{1pt}{1pt}
\bottomrule
\end{tabular}
\begin{tablenotes}\tiny
\item ${\star}$ The initial conditions contain initial longitudinal position and velocity, $p_{x, \rm ter}$ is $1000 \ {\rm m}$ in Fig.~\ref{fig:merging_scenario}.
\end{tablenotes} 
\end{table}
\begin{table}[!t]\tiny
\centering  
\caption{Comparison of the Performance of Three Planners in Stochastic Simulations}  
\label{Table_stochastic_comparison}  
\setlength{\tabcolsep}{1mm}
\begin{tabular}{ccccccccccc}  
\toprule  
\multicolumn{2}{c}{\underline{\textbf{Planner \& success rate}}}  & \multicolumn{3}{c}{\underline{\textbf{Merging approach of the EV}}} & \multicolumn{2}{c}{\underline{${\rm min}(d_{\rm ev}^{\rm sv0})$}} & \multicolumn{2}{c}{\underline{${\rm min}(d_{\rm ev}^{\rm sv1})$}} & \multicolumn{2}{c}{\underline{${\rm max}(|a^{\rm ev}|)$}} \\
\multicolumn{1}{c}{Name} & \multicolumn{1}{c}{Success rate} & \multicolumn{1}{c}{Ahead} & \multicolumn{1}{c}{Between} & \multicolumn{1}{c}{After} & \multicolumn{1}{c}{Mean} & \multicolumn{1}{c}{STD} & \multicolumn{1}{c}{Mean} & \multicolumn{1}{c}{STD} & \multicolumn{1}{c}{Mean} & \multicolumn{1}{c}{STD}  \\
\midrule
Proposed  & $100\%$ & $300/300$ & $0/300$ & $0/300$ & $4.06$ & $0.08$ & $44.9$ & $0.44$ & $1.28$ & $0.02$\\ \specialrule{0em}{1pt}{1pt}
DMPC & $92\%$ & $275/300$ & $0/300$ & $0/300$ & $0.15$ & $0.04$ & $43.9$ & $0.87$ & $2.50$ & $0.00$\\ \specialrule{0em}{1pt}{1pt}
RMPC & $100\%$ & $0/300$ & $0/300$ & $300/300$ & $82.3$ & $1.96$ & $30.3$ & $1.80$ & $4.82$ & $0.00$ \\ \specialrule{0em}{1pt}{1pt}
\bottomrule
\end{tabular}
\begin{tablenotes}\tiny
\item ${\star}$ 'Ahead' means the EV merges ahead of SV0; 'Between' means merging between SV0 and SV1; 'After' means merging after SV1. The unit of ${\rm min}(d_{\rm ev}^{\rm sv0})$ and ${\rm min}(d_{\rm ev}^{\rm sv1})$ is ${\rm m}$, and that of ${\rm max}(|a^{\rm ev}|)$ is ${\rm m/s^2}$.
\end{tablenotes} 
\end{table}
\begin{table}[!t]\tiny
\centering
\caption{Computational Performance}  
\label{Table_computation} 
\begin{tabular}{ccccc}
\toprule
\textbf{Method} & \textbf{Proposed} & \textbf{DMPC} & \textbf{RMPC} & \textbf{IA-BMPC} \\
\midrule
Mean $\pm$ Std. & $0.16 \pm 0.02 \ {\rm s}$ & $0.05 \pm 0.03\ {\rm s}$ & $0.16 \pm 0.02 \ {\rm s}$ & $0.50 \pm 0.16 \ {\rm s}$ \\ \specialrule{0em}{1pt}{1pt}
\bottomrule
\end{tabular}
\end{table}
\begin{figure}[!t]
    \centering
    \subfigure{\includegraphics[width=\columnwidth]{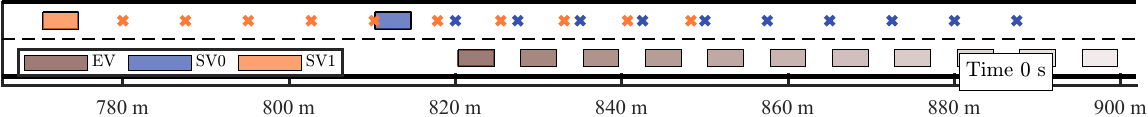}}
    \vfil
    \subfigure{\includegraphics[width=\columnwidth]{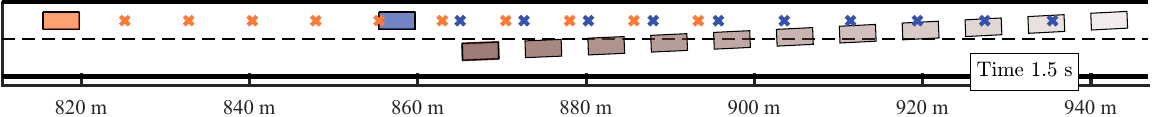}}
    \vfil
    \subfigure{\includegraphics[width=\columnwidth]{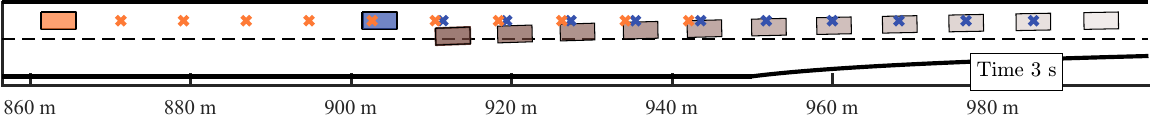}}
    \vfil
    \subfigure{\includegraphics[width=\columnwidth]{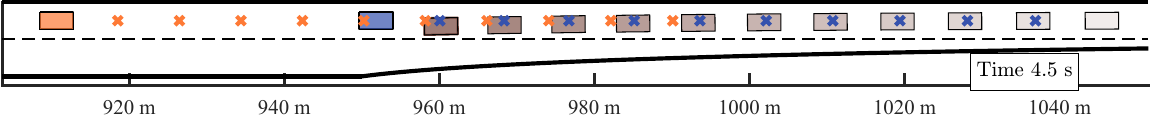}}
    \centering
    \caption{The merging process of the EV planned by the \textbf{proposed method}. The minimum distance between EV and SV0 in the same lane is $\textbf{3.97} \ {\rm m}$.}
    \label{fig:proposed method_Merging_Process}
    \end{figure}
    \begin{figure}[!t]
    \centering
    \subfigure{\includegraphics[width=\columnwidth]{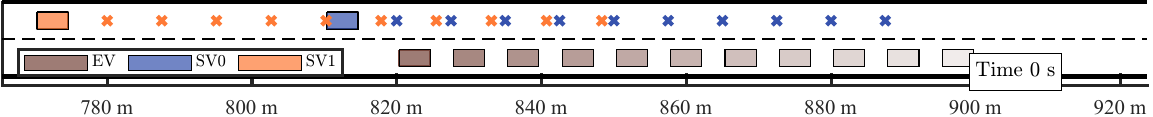}}
    \vfil
    \subfigure{\includegraphics[width=\columnwidth]{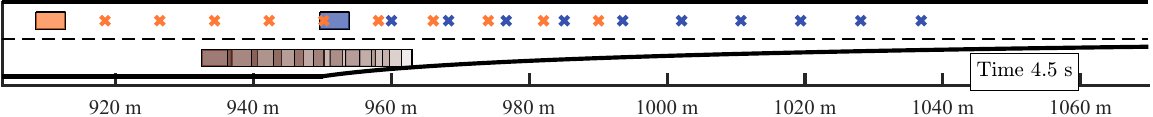}}
    \vfil
    \subfigure{\includegraphics[width=\columnwidth]{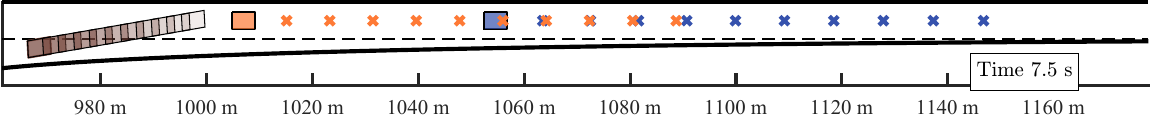}}
    \vfil
    \subfigure{\includegraphics[width=\columnwidth]{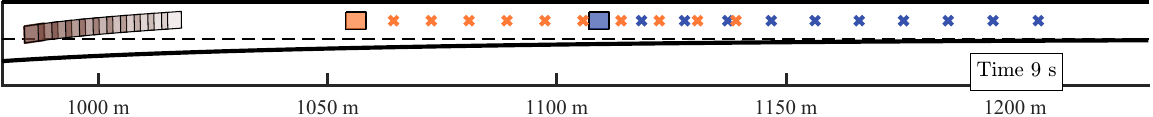}}
    \centering
    \caption{The merging process of the EV planned by \textbf{RMPC}. The minimum distance between EV and SV1 in the same lane is $\textbf{34.3} \ {\rm m}$.}
    \label{fig:RMPC_Merging_Process}
    \end{figure}
    \begin{figure}[!h]
    \centering
    \subfigure{\includegraphics[width=\columnwidth]{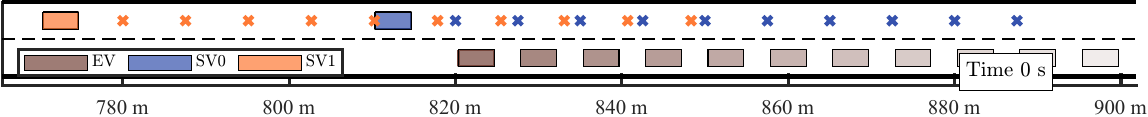}}
    \vfil
    \subfigure{\includegraphics[width=\columnwidth]{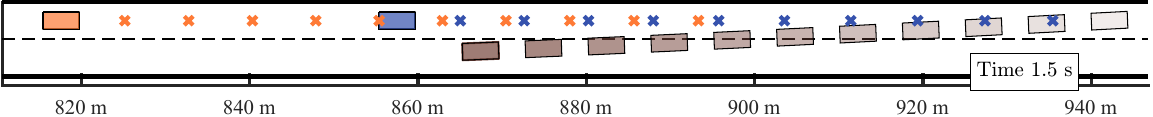}}
    \vfil
    \subfigure{\includegraphics[width=\columnwidth]{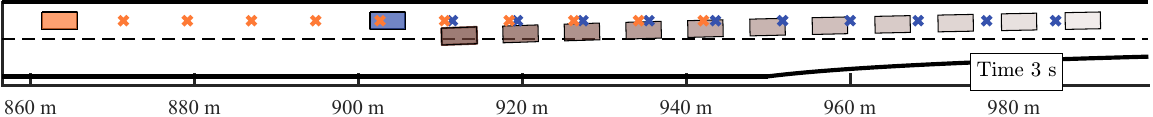}}
    \vfil
    \subfigure{\includegraphics[width=\columnwidth]{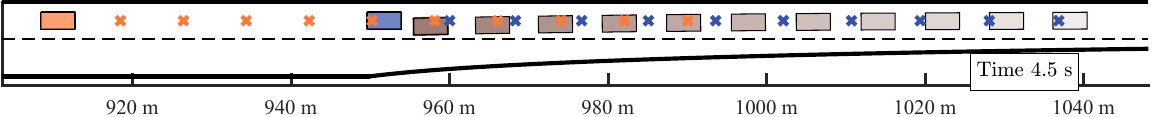}}
    \centering
    \caption{The merging process of the EV planned by \textbf{DMPC}. The minimum distance between EV and SV0 in the same lane is $\textbf{0.17} \ {\rm m}$.}
    \label{fig:DMPC_Merging_Process}
    \end{figure}
    \begin{figure}[!t]
    {\includegraphics[width=\columnwidth]{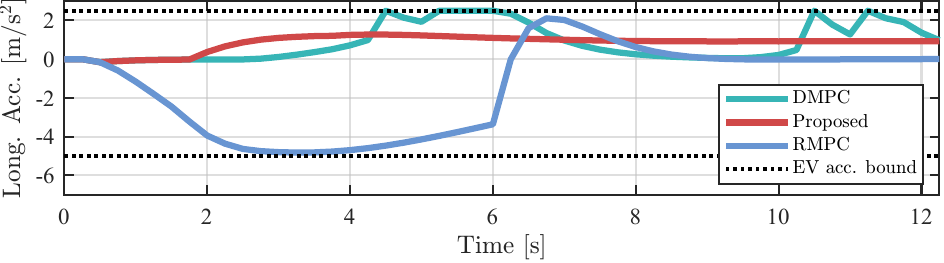}}
    \caption{The planned longitudinal acceleration of the EV by three methods corresponding to the results in Figs.~\ref{fig:proposed method_Merging_Process}--\ref{fig:DMPC_Merging_Process}.}
    \label{fig:SIM1_Single_Case_y_a_d}
    \end{figure}

\subsection{Comparison with Deterministic and Robust Approaches} \label{sec: compare with R and D}
In the first case study, an RMPC and a DMPC are implemented by replacing $\hat{\mathbb{U}}_t^s$ in \eqref{eq: reachable set prediction} with $\mathbb{U}^s$ and ${\bm 0}$, respectively. The three planners are first compared in a challenging merging scenario to get insights into the performance of each method. Then $300$ Monte-Carlo simulations are run in random environments to compare them in terms of statistical performance. To make the scenario challenging for the EV, the SV0 is designed to suddenly accelerate with random accelerations when encountering the EV around $p_{x,{\rm ter}}$ in Fig.~\ref{fig:merging_scenario}.  The set $\mathcal{I}_0^s$ is a minor, yet non-empty initial information set. This means that the EV starts to estimate the SVs' uncertainties without any prior knowledge. The three planners are implemented sequentially in the same scenario, and details of the merging process of the EV  are presented in Figs.~\ref{fig:proposed method_Merging_Process}--\ref{fig:DMPC_Merging_Process}. Fig.~\ref{fig:proposed method_Merging_Process} shows that the EV with the proposed method can merge in front of SV0. In contrast, the EV with RMPC in Fig.~\ref{fig:RMPC_Merging_Process} starts to merge much later as the EV needs to decelerate until the SVs pass to initiate the merging. Finally, Fig.~\ref{fig:DMPC_Merging_Process} shows that the EV with DMPC also merges in front of the SV0, but the distance between the EV and SV0 is smaller. Fig.~\ref{fig:SIM1_Single_Case_y_a_d} shows that the acceleration with the proposed method is smoother during the merging process, whereas the accelerations with the other methods touch the bound. 

The methods are further compared in $300$ Monte-Carlo simulations with the same simulation parameters as in the previous case. The methods are compared by counting the success rate of merging, where a successful case means that the EV merges to lane 2 without any collision with the SVs and the road boundary. Among the successful cases of each method, the merging approach, i.e., merging in front of the SV0, merging between SV0 and SV1, and merging after SV1, is also compared. In addition, we use three random variables, the minimum polytope distance between the EV and SV0 of each simulation when the EV is in lane 2 (${\rm min}(d_{\rm ev}^{\rm sv0})$), the minimum polytope distance between the EV and SV1 of each simulation when the EV is in lane 2 (${\rm min}(d_{\rm ev}^{\rm sv1})$), and the maximum absolute acceleration of the EV of each simulation (${\rm max}(|a^{\rm ev}|)$), to compare the three methods. The results are summarized in Table~\ref{Table_stochastic_comparison}. It is seen that the success rate of the proposed method and RMPC is $100\%$, while the rate of DMPC is lower. Among the successful cases, the proposed method always makes the EV merge in front of the SV0, and the RMPC makes the EV merge after SV1. The comparisons reflect that the proposed method is safer than the DMPC and less conservative than the RMPC, and can reduce the control input magnitudes of the EV compared with both DMPC and RMPC. 

\subsection{Comparison with Interaction-Aware Branch MPC}\label{sec: compare with branch MPC}
\begin{figure}[!t]
\centering
\subfigure{\includegraphics[width=\columnwidth]{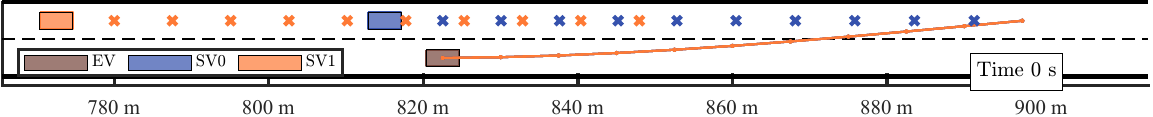}}
\vfil
\subfigure{\includegraphics[width=\columnwidth]{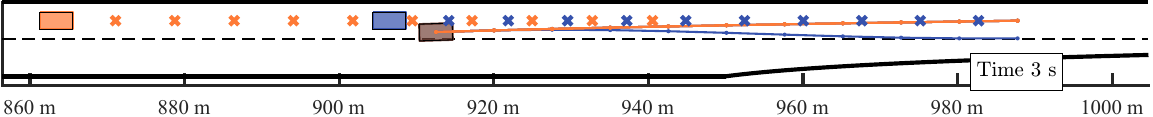}}
\vfil
\subfigure{\includegraphics[width=\columnwidth]{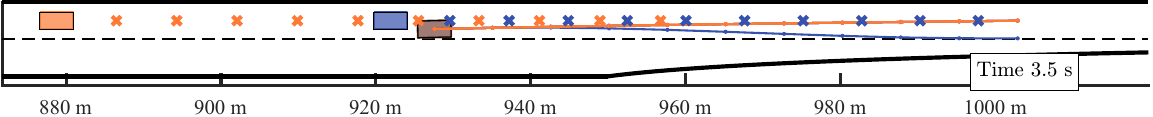}}
\vfil
\subfigure{\includegraphics[width=\columnwidth]{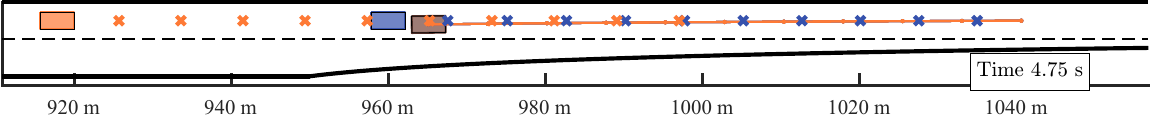}}
\centering
\caption{The merging process of the EV with IA-BMPC. The minimum distance between EV and SV0 in the same lane is $\textbf{0.74} \ {\rm m}$ (compared with $\textbf{2.66} \ {\rm m}$ by the proposed method).}
\label{fig:BMPC_Merging_Process}
\end{figure}
\begin{figure}[!t]
{\includegraphics[width=\columnwidth]{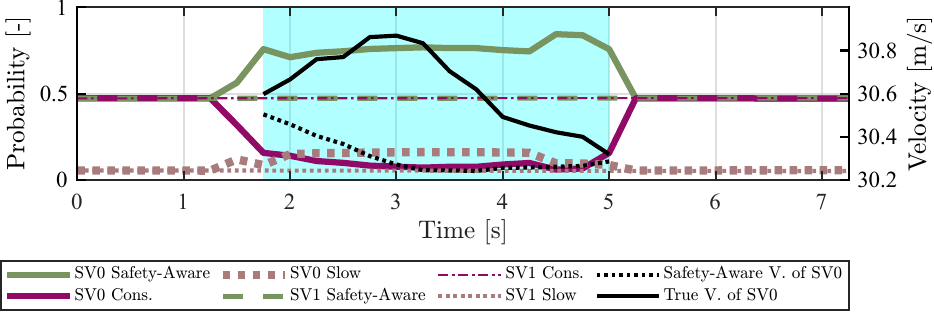}}
\caption{The predicted probability of each mode of each SV, and the predicted reference speed and the true speed of SV0 in the IA-BMPC.}
\label{fig:SIM4_Single_Case_general}
\end{figure}
The second case study compares the proposed method with an interaction-aware branch MPC (IA-BMPC), as highlighted in Section~\ref{sec: related workd}, which has shown efficiency in solving motion-planning problems in autonomous driving. The branch MPC can be realized and implemented in different ways \cite{fors2022resilient}, \cite{chen2022interactive}, \cite{wang2023interaction}, \cite{oliveira2023interaction}, while the key idea is to estimate the multi-modal uncertain intentions of the SVs using a motion-prediction model. Following the multi-modal uncertainty predictions, the motion-planning method decides branching points and then computes multiple control policies associated with the multi-modality of the SVs \cite[Section 3.3.1]{soman2023learning}. If the SVs are predicted with an interaction-aware method, the branch MPC refers to IA-BMPC. In this paper, the uncertain intentions of SVs with the IA-BMPC are described by three modes, they are (i) safety-aware mode, (ii) constant-speed mode, and (iii) slowdown mode. The safety-aware mode optimizes the reference velocity of the SV subject to the collision-avoidance constraint with the EV, and the reference velocity is computed by an interaction-aware prediction model adapted from \cite{zhou2022interactionnew}, \cite{lefkopoulos2021interaction}. The constant-speed mode tracks the current velocity, and the slowdown mode tracks a low velocity. Following the reference velocity, the trajectories of the SVs are predicted by a linear state-feedback controller, and the probability of each mode of the SVs is predicted considering the maneuver cost and collision risk with the EV. The predicted trajectories, as well as the probabilities of all modes of the SVs, are finally substituted into a branch MPC framework to optimize the policies of the EV. The implementation details are found in the published code. 

The IA-BMPC and the proposed method are implemented in the same scenario, where the merging process of the EV planned by the IA-BMPC is shown in Fig.~\ref{fig:BMPC_Merging_Process} (the merging process planned by the proposed method is similar to that in Fig.~\ref{fig:proposed method_Merging_Process}), with the predicted probabilities of each mode of the IA-BMPC are shown in Fig.~\ref{fig:SIM4_Single_Case_general}. It is seen in Fig.~\ref{fig:BMPC_Merging_Process} that the EV with IA-BMPC successfully merges in front of SV0, and the IA-BMPC optimizes different policies adapted to the multi-modal predictions of the SVs. However, the distance between the EV and SV0 with the IA-BMPC is smaller than that with the proposed method. This is because, as shown in Fig.~\ref{fig:SIM4_Single_Case_general}, when the EV tries to merge in front of the SV0, the prediction model predicts that the safety-aware mode of the SV0 has the highest probability. However, the planned maneuver and the predicted maneuver of SV0 have deviations, as shown in Fig.~\ref{fig:SIM4_Single_Case_general}, where the true velocity of the SV0 does not reduce as fast as predicted. This is because the IA-BMPC cannot accurately capture the uncertainties of the SVs, which reduces the robustness of the method, particularly in emergent scenarios where the EV and SVs are very close to each other. This indicates that the performance of the intention-prediction based method can be affected when the prediction model cannot accurately infer the intention of SVs, while the proposed method maintains robustness without knowing the intention of SVs. In addition, the proposed method also outperforms the IA-BMPC in terms of computational performance, as shown in Table~\ref{Table_computation}.

\subsection{Convergence Analysis}\label{sec: convergence}
\begin{figure}[!t]
\centering
\subfigure[ ]{\includegraphics[width=\columnwidth]{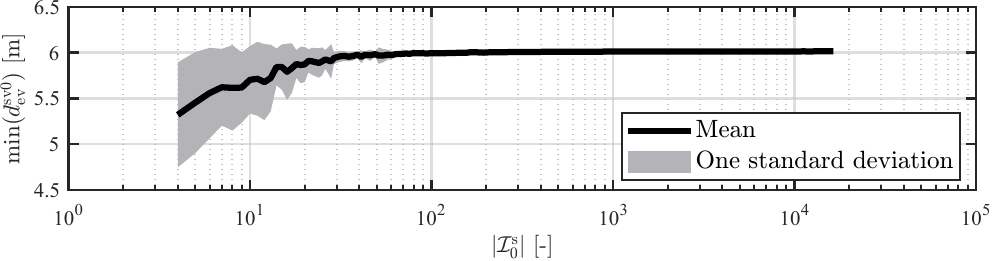}}
\vfil
\subfigure[ ]{\includegraphics[width=\columnwidth]{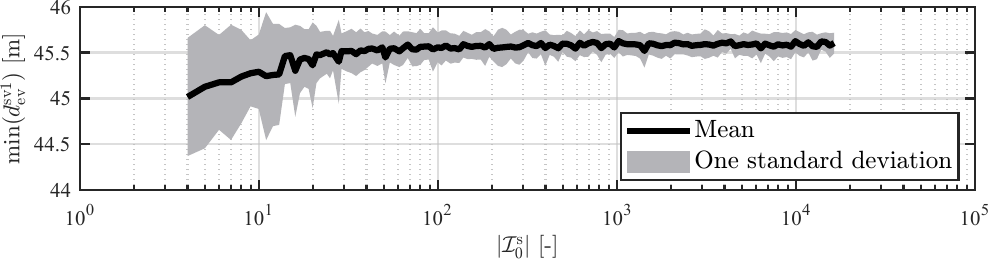}}
\vfil
\subfigure[ ]{\includegraphics[width=\columnwidth]{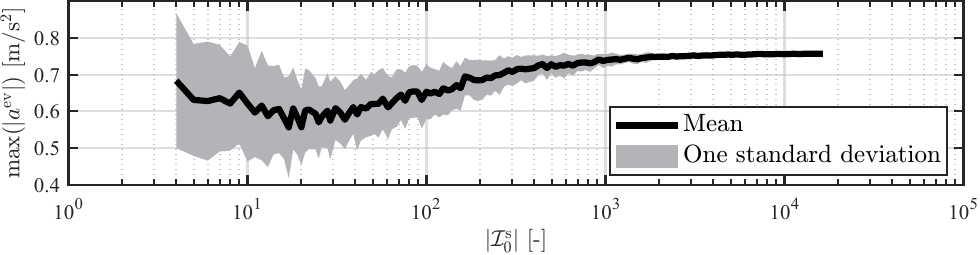}}
\centering
\caption{Statistical performance of the proposed method by changing $|\mathcal{I}_0^s|$. (a) The minimum distance between the EV and SV0. (b) The minimum distance between the EV and SV1. (c) The absolute maximum acceleration of the EV. Note that the success rate of EV's merging behavior with each $|\mathcal{I}_0^s|$ in the simulations is always $100\%$.}
\label{fig:SIM3_Change_Sample_Size}
\end{figure}
The third case study analyzes the impact of the information set on the performance of the proposed method. The initial information set $\mathcal{I}_0^s$ was designed containing the samples from the distribution of accelerations of the SVs, where the parameter $|\mathcal{I}_0^s|$ (the cardinality of $\mathcal{I}^s_0$) increases from $4$ to $16384$. With each $|\mathcal{I}_0^s|$ the simulation was run $50$ times to count the mean and standard deviation of the random variables ${\rm min}(d_{\rm ev}^{\rm sv0})$, ${\rm min}(d_{\rm ev}^{\rm sv1})$, and ${\rm max}(|a^{\rm ev}|)$. The statistical results are shown in Fig.~\ref{fig:SIM3_Change_Sample_Size}, which shows that when $|\mathcal{I}_0^s|$ increases, these variables converge to a stable value. This means that the robustness of the method is increased when the information set is richer. Note that the elements in $\mathcal{I}_0^s$ and $\mathcal{I}_t^s$ are generated in the same way, so Fig.~\ref{fig:SIM3_Change_Sample_Size} also reflects how these variables converge during the online iterations with $t$. 

\subsection{Computational Time}
The calculations were performed on a standard laptop with an Intel i7-10750H CPU, 32.0 GB RAM running Ubuntu 22.04 LTS and Python 3.10.12. The computation time of each method is summarized in Table~\ref{Table_computation}. These results were recorded from $20$ random simulations with each MPC algorithm running $40$ steps in each simulation. 

\section{Conclusion} \label{Conclusion}
This paper studied resilient decision-making and motion-planning of an autonomous ego vehicle in forced merging scenarios based on estimating the acceleration bounds of the surrounding vehicles. The estimated acceleration bounds were used to compute the forward occupancy of the surrounding vehicles over a horizon, which was integrated into a decision-making strategy to compute the safe reference state of the ego vehicle. Then, a robust MPC controller was designed to plan the trajectory by tracking the reference state subjected to safety constraints with obstacle occupancy. Simulation results show that: (1) The method is safer than a deterministic method and less conservative than a robust method; (2) The method does not need to estimate the intentions of surrounding vehicles while still maintaining desired robustness in uncertain environments; (3) The performance is convergent when more acceleration information of surrounding vehicles is observed. 
\addtolength{\textheight}{-12cm} 
\bibliography{IEEEabrv,mybib}{}
\bibliographystyle{IEEEtran}
\end{document}